\documentclass[sigconf]{acmart}

\copyrightyear{2026}
\acmYear{2026}
\setcopyright{cc}
\setcctype{by}
\acmConference[ACM Sustainability Week Companion '26]{ACM Sustainability Week 2026}{June 22--25, 2026}{Banff, AB, Canada}
\acmBooktitle{ACM Sustainability Week 2026 (ACM Sustainability Week Companion '26), June 22--25, 2026, Banff, AB, Canada}
\acmDOI{10.1145/3765611.3815359}
\acmISBN{979-8-4007-2199-1/2026/06}

\usepackage{algorithm}
\usepackage{algorithmic}
\usepackage{colortbl}
\usepackage{booktabs}
\usepackage{siunitx}
\usepackage{textcomp}

\DeclareSIUnit{\EUR}{\text{\texteuro}}

\begin{document}

\title[]{Generalizing HVAC Control With Domain Randomized Reinforcement Learning}

\author{Pablo Boitel}
\email{pablo.boitel.1@ens.etsmtl.ca}
\affiliation{%
  \institution{École de technologie supérieure (ÉTS)}
  \city{Montréal}
  \state{Québec}
  \country{Canada}
}

\author{Kun Zhang}
\email{kun.zhang@etsmtl.ca}
\affiliation{%
  \institution{École de technologie supérieure (ÉTS)}
  \city{Montréal}
  \state{Québec}
  \country{Canada}
}

\begin{abstract}
Deploying advanced HVAC (Heating, Ventilation and Air Conditioning) controllers at scale remains difficult because performance often depends on accurate building models or per-site retuning. We propose \textsc{NOMAD-RL} (Neural Online Meta-Adaptation for Dynamics), a general-purpose Reinforcement Learning (RL) controller designed to transfer across heterogeneous thermal zones through a universal, non-invasive thermostat interface. The controller acts on temperature setpoints from zone measurements and forecasts, while a recurrent policy supports online adaptation under partial observability.

Our main contribution is an adaptive domain randomization scheme based on physics-informed normalizing flows, which models correlated and multimodal distributions of thermal-zone parameters while maintaining physical plausibility and controllability. This produces a realistic and progressively adaptive training curriculum that improves transfer across buildings.
We evaluate \textsc{NOMAD-RL} against a constant-setpoint PID controller, RL without domain randomization, and MPC in single- and multi-zone settings. \textsc{NOMAD-RL} consistently outperforms the PID and non-randomized RL baselines, and approaches the performance of a well-tuned MPC, especially in the more challenging multi-zone case. These results highlight the potential of adaptive, physics-informed domain randomization for robust and transferable HVAC control.
\end{abstract}




\maketitle

\section{Introduction}

Buildings account for $30\%$ of global final energy use and $26\%$ of energy-related $CO_2$ emissions \cite{noauthor_buildings_nodate}, making HVAC control a promising lever for reducing both operating costs and emissions.

Model Predictive Control (MPC) remains the reference approach because it can explicitly balance comfort and energy use under operational constraints. However, its large-scale deployment is limited by the effort required to derive, calibrate, and maintain accurate models, especially beyond simple single-zone settings \cite{drgona_all_2020,nagy_ten_2023}. Reinforcement Learning (RL) has therefore emerged as an alternative. Yet RL offers little advantage when it also relies on highly calibrated simulators: MPC depends on models explicitly, whereas RL depends on them implicitly through training environments.

This motivates a different objective for RL in HVAC control: learn controllers that transfer across heterogeneous thermal zones. The key intuition is that many physically different zones, exhibit similar closed-loop behavior when controlled through a thermostat interface. A controller trained across varied zones, while retaining memory of past dynamics and adapting online, can therefore learn transferable strategies rather than zone-specific heuristics.

We propose \textsc{NOMAD-RL} (\textbf{N}eural \textbf{O}nline \textbf{M}eta-\textbf{A}daptation for \textbf{D}ynamics), a general-purpose RL controller designed for this setting. \textsc{NOMAD-RL} acts through a universal thermostat interface and uses recurrent policies to handle partial observability and online adaptation. To promote transfer, training relies on adaptive domain randomization over thermal-zone dynamics. Rather than sampling simulator parameters independently, we model their joint distribution with neural spline flows and constrain it with physics-informed regularization so that training remains both diverse and physically plausible.

Our contributions are fourfold. First, we introduce a flow-based adaptive domain randomization scheme that captures correlated and multimodal distributions of thermal-zone parameters. Second, we enforce physical plausibility and controllability through physics-informed regularization. Third, we propose a stable training objective based on a context-normalized reward and a Boltzmann target with gradient-refined candidates, improving sample efficiency in high dimension. Fourth, we combine this curriculum with a recurrent RL controller that exploits interaction history and forecasts to support rapid adaptation under partial observability.

Experiments in both single-zone and multi-zone settings show that \textsc{NOMAD-RL} consistently outperforms constant-setpoint and non-randomized RL baselines, while approaching the performance of a well-tuned MPC. These results suggest that adaptive, physics-informed domain randomization can make RL a practical path toward robust and transferable HVAC control across diverse buildings.

\begin{figure*}[h]
    \centering
    \includegraphics[width=0.8\linewidth]{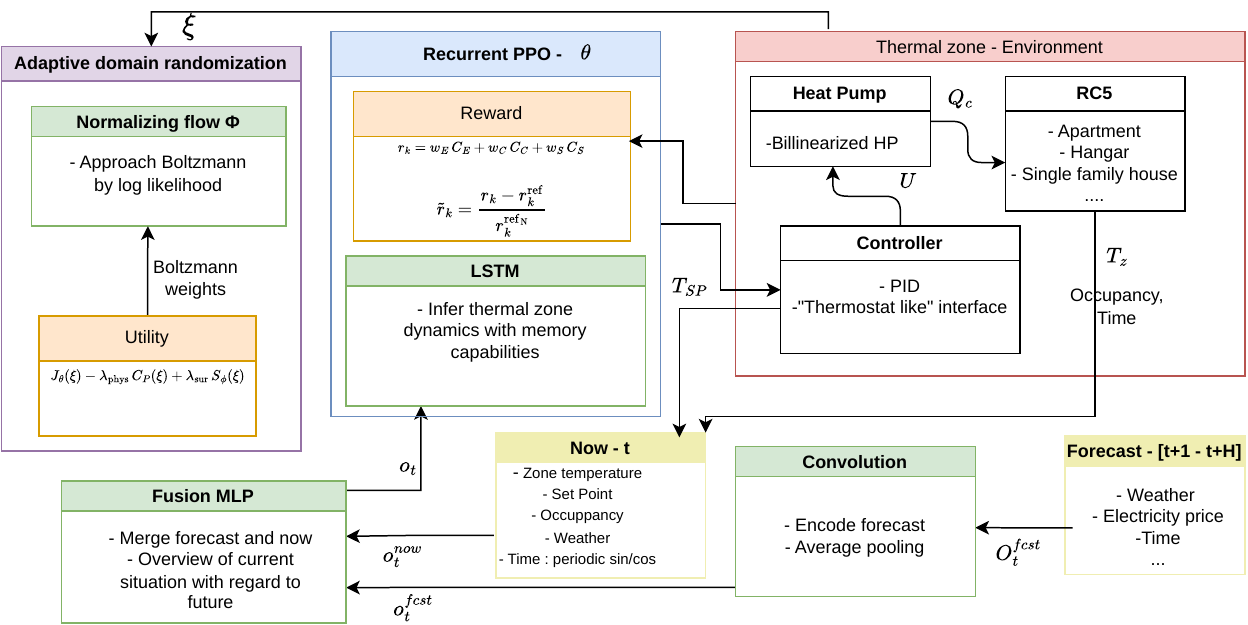}
    \caption{Architecture of \textbf{NOMAD-RL}}
    \label{fig:schema_architecture}
\end{figure*}

\section{Method}

We consider a general and practical control setting in which the agent interacts with the building through a standard thermostat interface. Rather than directly controlling the heating plant, the controller observes zone temperature and power, adjusts the temperature setpoint. This supervisory control approach improves safety, since actions can be constrained within a reasonable comfort range, preserves interpretability, and avoids learning low-level actuation already handled by existing PID loops. It also promotes transfer across buildings, as closed-loop thermostat-level behavior tends to be more consistent than plant dynamics. See Fig.\ref{fig:schema_architecture} for an overview of the architecture of \textbf{NOMAD-RL} and details are given below.

\subsection{Thermal zone modeling}

In this work, we use a generic thermal-zone simulator based on a fifth-order resistance-capacitance (RC5) model coupled with a hydronic heat pump and a low-level PID controller; following \cite{arroyo_reinforced_2022}. The model is implemented in JAX to support efficient simulation and automatic differentiation.


The parameters include those of building envelope, internal mass, heat pump characteristics, and PID gains, resulting in a 25-dimensional domain space. This compact but expressive setup provides a convenient testbed for evaluating transfer, robustness, and adaptive domain randomization, while still allowing strong MPC baselines to be computed on the nominal model.

\subsection{Reinforcement learning controller}

We formulate HVAC control as a partially observable sequential decision-making problem (POMDP). Since the controller does not observe the full thermal states of the building, it must act from limited measurements and delayed effects. This motivates a recurrent RL policy to infer hidden dynamics from interaction history. At each step, the agent observes the current zone temperature, power consumption, comfort bounds, outdoor conditions, occupancy, electricity price, and time features, together with short-term forecasts of the main exogenous inputs. Rather than directly controlling the heating plant, the policy outputs an incremental update to the thermostat setpoint, clipped to a safe comfort range. This preserves interpretability, simplifies deployment, and keeps low-level actuation handled by the existing PID loop.

Forecast trajectories are encoded with a temporal convolution module and fused with current measurements into a compact latent representation. This representation is then passed to a recurrent LSTM-based actor--critic policy, providing the required memory to adapt online to partially observed and changing dynamics. We train the controller with Recurrent PPO, using the standard actor--critic implementation from the SB3-Contrib library.

The reward reflects the operational HVAC objective: minimizing electricity cost, maintaining comfort and avoiding excessive actuator saturation. To make training comparable across heterogeneous buildings and operating conditions, we use a normalized reward relative to a fixed-setpoint PID reference rollout computed at the start of each episode. This replay-based normalization differs from the standard online RL setting, where trajectories are stochastic, non-reproducible, and cannot be exactly replayed. It therefore provides a stable learning signal under domain randomization and measures improvement over a simple but realistic reference controller.

\subsection{Domain randomization for RL training}

To promote transfer across heterogeneous thermal zones, we train the policy under domain randomization. At the beginning of each episode, all the simulator parameters $\xi \in \mathbb{R}^{25}$ are sampled from a distribution $p_\phi(\xi)$, while the policy does not observe $\xi$ directly. Training therefore optimizes expected return over a family of environments rather than a single nominal model.

The main challenge is to adapt $p_\phi$ so that it generates informative and realistic environments: if the distribution is too narrow, the policy overfits; if it is too broad, training becomes unstable. Building upon previous work \cite{openai_solving_2019, tiboni_domain_2024, curtis_flow-based_nodate}, we address this challenge with an adaptive domain randomization scheme based on normalizing flows.

We represent $p_\phi(\xi)$ with a neural spline flow, which can model correlated, multimodal, and high-dimensional parameter distributions. Rather than optimizing the flow directly through noisy gradients, we construct an empirical Boltzmann target over candidate environments and fit the flow to it by weighted maximum likelihood, yielding a more stable update.

To remain effective in high-dimensional settings, candidate parameters are first sampled uniformly over the admissible domain, then refined through a few gradient steps. Each random draw thus becomes a short local search trajectory, which improves coverage of the parameter space.
Each candidate is evaluated using three components: the return under the current policy $J_\theta(\xi)$, a physical plausibility cost $C_P(\xi)$, and a surprise bonus $\log p_\phi(\xi)$ that promotes exploration. The resulting utility $U(\xi)$, which measures the relevance of a parameter vector (higher is better), is defined as
\begin{equation}
U(\xi) = J_\theta(\xi) - \lambda_{\mathrm{phys}} C_P(\xi) + \lambda_{\mathrm{sur}} \bigl(-\log p_\phi(\xi)\bigr)
\label{eq:utility_nomad}
\end{equation}

This utility yields the Boltzmann weights, i.e., the analytical solution of the associated expectation-maximization problem:
\begin{equation}
w_i = \mathrm{softmax}\left(\frac{U(\xi_i)}{\tau}\right)_i
\label{eq:boltz_weights_nomad}
\end{equation}
where $\tau$ controls how concentrated the distribution is.

\begin{figure}[h]
    \centering
    \includegraphics[width=0.9\linewidth]{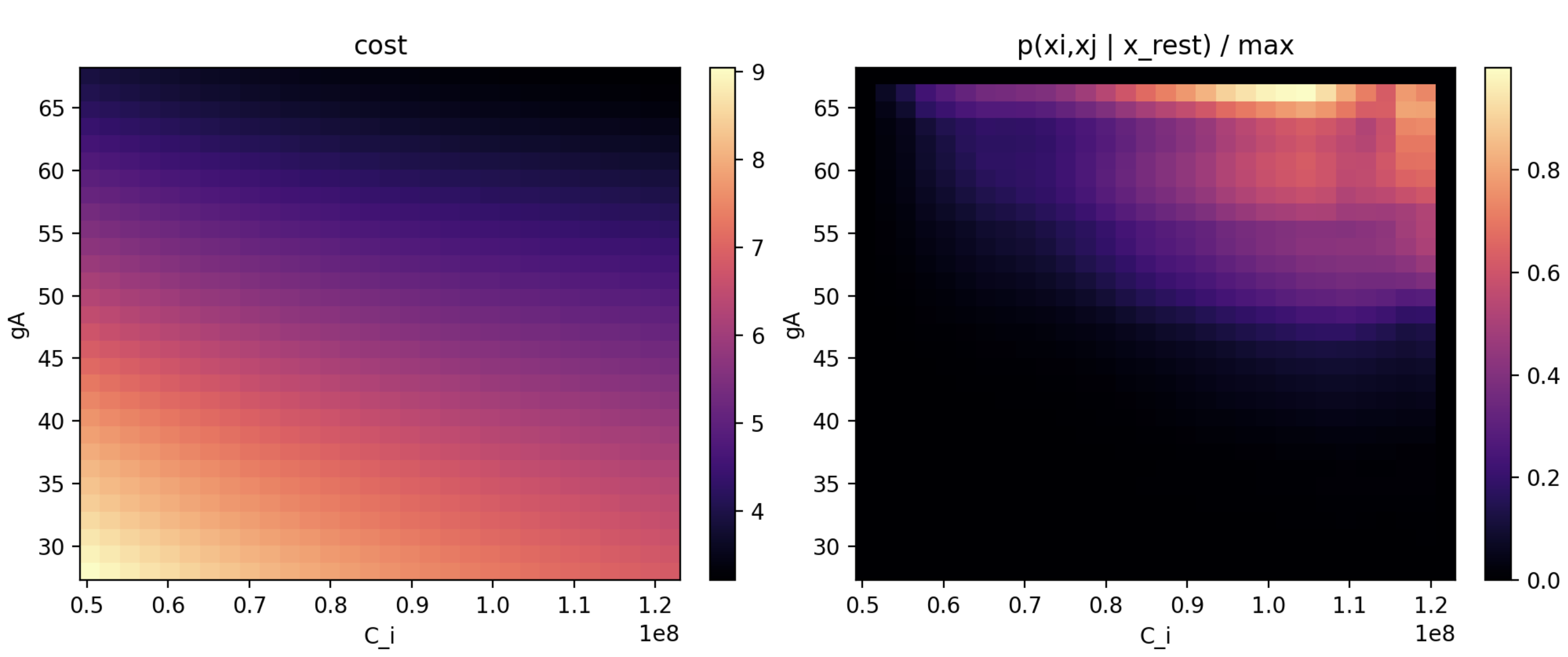}
    \caption{2D slice of the learned parameter-flow distribution in the $(gA, C_i)$ space over the target Boltzmann distribution (left) and their correlation (right)}
    \label{fig:expanded}
\end{figure}

The physical cost $C_P(\xi)$ constrains the curriculum to realistic and controllable regions of the parameter space by penalizing implausible heat-pump efficiencies, poor comfort regulation under a fixed-setpoint baseline controller, and excessive actuator saturation. Figure~\ref{fig:expanded} shows a 2D slice of the resulting learned distribution in the $(gA, C_i)$ space, i.e., equivalent solar area and internal capacitance.

The parameter distribution is then updated from the weighted candidate set using a Boltzmann-weighted negative log-likelihood with a KL trust region:
\begin{equation}
\min_{\phi}\quad
L(\phi)
=
-\sum_{i=1}^{N} w_i \log p_\phi(\xi_i)
+
\beta D_{\mathrm{KL}}\!\left(p_\phi\,\|\,p_{\phi_{\mathrm{old}}}\right)
\label{eq:flow_loss_nomad}
\end{equation}

This stabilizes successive updates while allowing the distribution to expand gradually as the policy improves. Overall, \textbf{NOMAD-RL} alternates between updating the recurrent policy on environments sampled from the current flow and updating the flow from refined candidates scored by utility, yielding an adaptive curriculum over realistic thermal-zone dynamics (see the Algorithm 1 for the steps).

\subsection{Training configuration}

Training was limited to a preliminary $6e6$-steps run on a single CPU/GPU, using Algorithm~\ref{alg:nomad}, with limited hyperparameter tuning.

\begin{figure*}[h]
    \centering
    \includegraphics[width=1.0\linewidth]{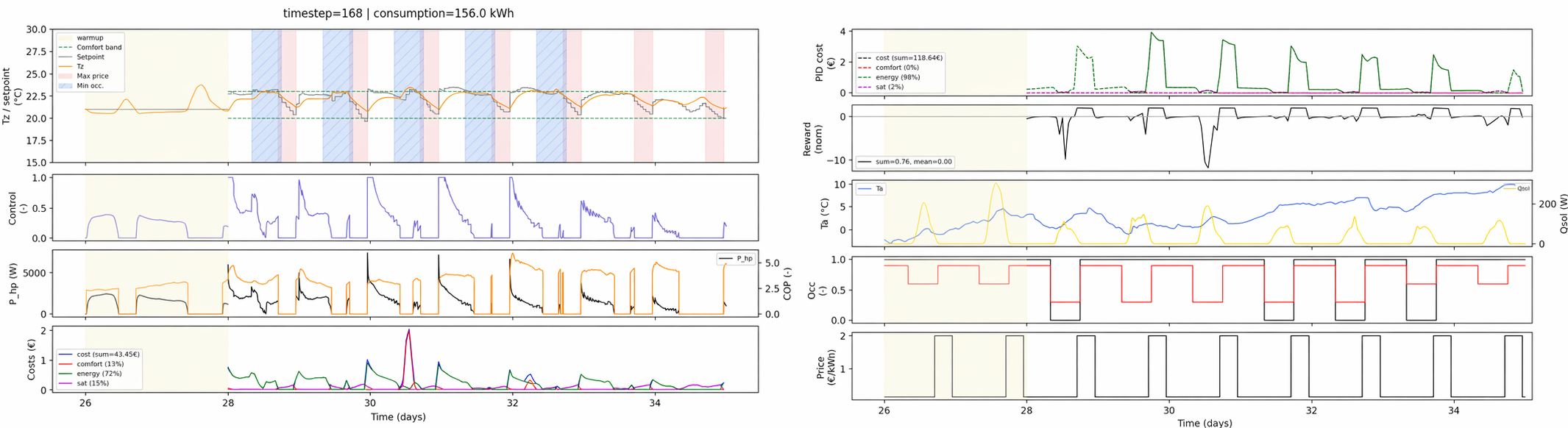}
    \caption{Behavior of the NOMAD-RL controller during training.}
    \label{fig:NOMAD_after_train}
\end{figure*}

Fig.~\ref{fig:NOMAD_after_train} illustrates a representative \textbf{NOMAD-RL} training episode in our tuned GYM environment, showing the full sequence of actions and observations. High-price events (red) are periods the controller must anticipate, where power should be minimized. Low-occupancy periods (blue) indicate times when temperature may temporarily drift outside comfort bounds without penalty. The episode starts with a warm-up phase (yellow). The figure also reports the constant-setpoint baseline, weather forecasts, power and COP trajectories, and an example occupancy profile.

\begin{algorithm}[h]
\caption{NOMAD-RL}
\label{alg:nomad}
\begin{algorithmic}[1]
\REQUIRE policy parameters $\theta$, flow parameters $\phi$
\FOR{$n = 1$ to $N$}
    \STATE \textsc{Train Policy}$(\theta, p_\phi)$
    \STATE $\mathcal{D} \leftarrow \textsc{Sample And Refine}(\mathcal{X}, U)$
    \STATE $w \leftarrow \textsc{Boltzmann Weights}(\mathcal{D}, U)$
    \STATE \textsc{Update Flow}$(\phi, \mathcal{D}, w)$
\ENDFOR
\end{algorithmic}
\end{algorithm}


\section{Results and evaluation of NOMAD-RL}

After training, we evaluate \textbf{NOMAD-RL} against a constant set-point PID controller, a recurrent PPO baseline trained without domain randomization, and MPC. \textbf{Costs are reported in euros, reflecting energy consumption, comfort, and saturation}. Percentages use $(C_{\mathrm{ref}} - C_{\mathrm{NOMAD}})/C_{\mathrm{ref}} \times 100$; negative values mean that \textbf{NOMAD-RL} is more costly than the reference.

\subsection{Single-zone setting}

We first compare \textbf{NOMAD-RL} with a recurrent PPO agent trained on a single fixed environment. Both policies are evaluated on the nominal zone and under randomized dynamics to assess robustness and out-of-domain generalization. We also compare them with MPC and a constant set-point PID controller on a representative sampled zone. In this setting, MPC has access to the exact thermal-zone model and serves as a near-global optimum reference.

\begin{figure}[h]
    \centering
    \includegraphics[width=1.05\linewidth]{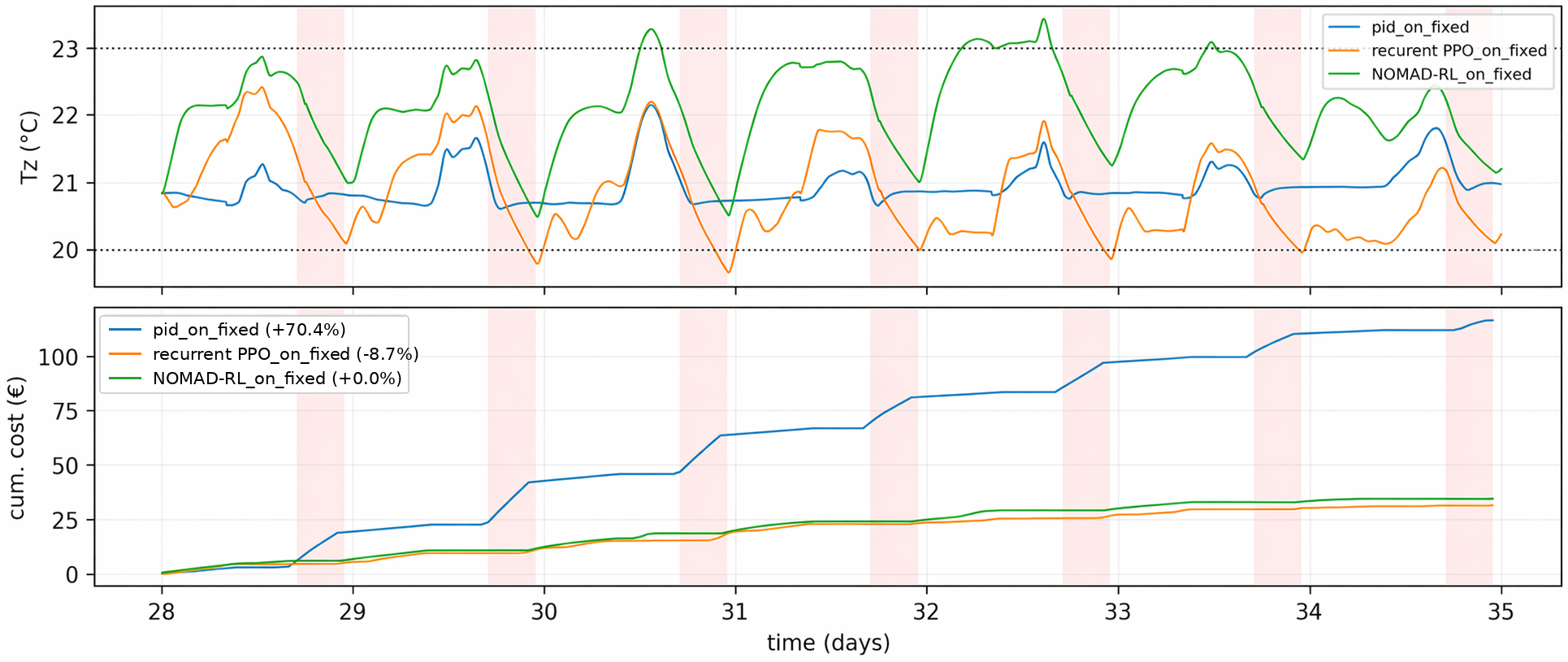}
    \caption{NOMAD-RL vs. recurrent PPO without domain randomization on the nominal thermal zone.}
    \label{fig:RL_results_double}
\end{figure}

\begin{figure}[h]
    \centering
    \includegraphics[width=1.05\linewidth]{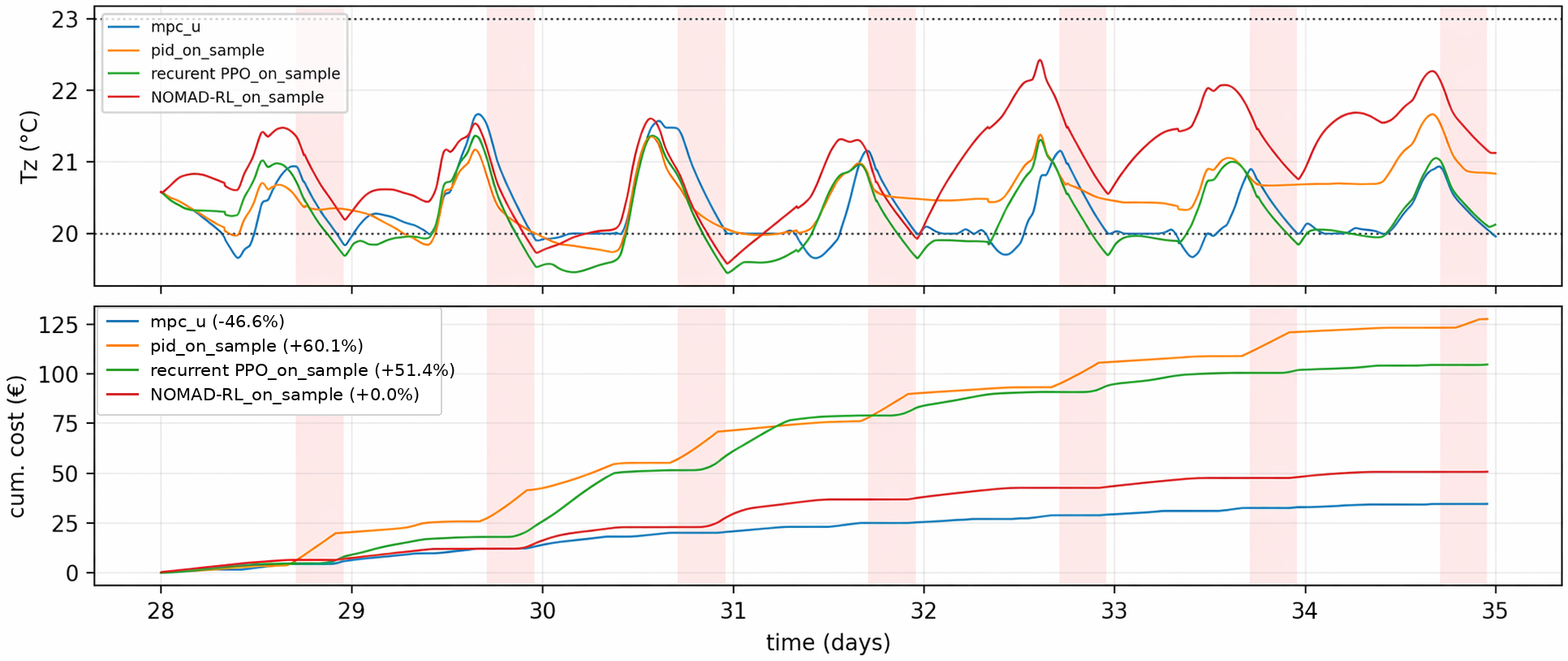}
    \caption{NOMAD-RL vs. recurrent PPO without domain randomization vs. MPC on a sampled thermal zone.}
    \label{fig:RL_results_sample}
\end{figure}

As shown in Fig.~\ref{fig:RL_results_double}, both RL controllers clearly outperform the constant set-point PID baseline. \textbf{NOMAD-RL} reduces cost by $70.4\%$ relative to PID, but is $8.7\%$ more costly than recurrent PPO on the nominal zone.

On the sampled zone with shifted dynamics (Fig.~\ref{fig:RL_results_sample}), \textbf{NOMAD-RL} reduces cost by $60.1\%$ relative to PID and $51.4\%$ relative to recurrent PPO, while being $46.6\%$ more costly than MPC. This is notable because MPC has access to the exact model and therefore approximates the global optimum in this setting.

\subsection{Multi-zone setting}

We compare \textbf{NOMAD-RL} with per-zone MPC and constant-setpoint PID on a 3-zone building with inter-zone thermal coupling, using the building-level cumulative cost aggregated over the three zones. 

\textbf{NOMAD-RL} remains effective in this more challenging setting: it reduces cost by $67.9\%$ relative to constant-setpoint PID and is $14.5\%$ more costly than MPC. The gap to MPC narrows as the problem becomes more complex, suggesting robustness in complex environments.

\section{Conclusion}

This work addresses a central challenge in scalable HVAC control: achieving robust performance across heterogeneous thermal zones without per-site modeling or retuning. We introduced \textbf{NOMAD-RL}, a general-purpose RL controller built on a universal thermostat interface that preserves safety, interpretability, and practical deployability. This supervisory approach is non-invasive and lightweight enough for embedded devices or connected thermostats.

Our main contribution is an adaptive domain randomization framework based on physics-informed normalizing flows. By learning multimodal, correlated distributions over thermal-zone parameters while enforcing physical plausibility and controllability, the framework generates a realistic and progressively expanding training curriculum. Combined with a recurrent policy leveraging past observations and forecasts, it yields transferable control policies capable of rapid online adaptation in a meta-learning setting.

Results support this approach. In single-zone settings, \textsc{NOMAD-RL} outperforms PID and non-randomized RL baselines, while remaining within $46.6\%$ of MPC on the sampled zone. In multi-zone settings, it reduces PID cost by $67.9\%$ and remains within $14.5\%$ of MPC. This suggests that \textsc{NOMAD-RL} provides a path toward generalizable HVAC control without accurate models.

Future work will focus on improving performance and real-world readiness through further training, hyperparameter tuning, expansion of the simulator and dataset pipeline, and validation on real buildings. We also plan to evaluate \textsc{NOMAD-RL} on HVAC benchmarks for generalization, and explore larger architectures, transformers, and flow matching for domain randomization.
The implementation used in this work is publicly available.\footnote{\url{https://github.com/paabL/NOMAD-RL}}

\bibliographystyle{ACM-Reference-Format}
\bibliography{references.bib}

\end{document}